%% file: main.tex
\documentclass[9pt, conference]{IEEEtran}

\usepackage{cite}
\usepackage{amsmath,amssymb,amsfonts}
\usepackage{graphicx}
\usepackage{textcomp}
\usepackage{booktabs} 
\usepackage{xcolor}
\usepackage[linesnumbered,ruled,vlined]{algorithm2e}

\begin{document}

\title{MT-Spike: A Multilayer Time-based Spiking Neuromorphic Architecture with Temporal Error Backpropagation}

\author{
\IEEEauthorblockN{Tao Liu\IEEEauthorrefmark{1}, Zihao Liu\IEEEauthorrefmark{1}, Fuhong Lin\IEEEauthorrefmark{2}, Yier Jin\IEEEauthorrefmark{3}, Gang Quan\IEEEauthorrefmark{1} and Wujie Wen\IEEEauthorrefmark{1}}

\IEEEauthorblockA{\IEEEauthorrefmark{1}\textit{Florida International University}, Miami, FL 33174, USA\\
\IEEEauthorrefmark{2}\textit{University of Science \& Technology Beijing}, Beijing 100083, CHINA\\
\IEEEauthorrefmark{3}\textit{University of Florida}, Gainesville, FL 32611, USA\\
\IEEEauthorrefmark{1}\{tliu023, zliu021, gang.quan, wwen\}@fiu.edu, \IEEEauthorrefmark{2}fhlin@ustb.edu.cn, \IEEEauthorrefmark{3}yier.jin@ece.ufl.edu}
}








\maketitle

\begin{abstract}

Modern deep learning enabled artificial neural networks, such as Deep Neural Network (DNN) and Convolutional Neural Network (CNN), have achieved a series of breaking records on a broad spectrum of recognition applications. However, the enormous computation and storage requirements associated with such deep and complex neural network models greatly challenge their implementations on resource-limited platforms. Time-based spiking neural network has recently emerged as a promising solution in Neuromorphic Computing System designs for achieving remarkable computing and power efficiency within a single chip. However, the relevant research activities have been narrowly concentrated on the biological plausibility and theoretical learning approaches, causing inefficient neural processing and impracticable multilayer extension thus significantly limitations on speed and accuracy when handling the realistic cognitive tasks. In this work, a practical multilayer time-based spiking neuromorphic architecture, namely ``MT-Spike'', is developed to fill this gap. 
With the proposed practical time-coding scheme, average delay response model, temporal error backpropagation algorithm and heuristic loss function, ``MT-Spike'' achieves more efficient neural processing through flexible neural model size reduction while offering very competitive classification accuracy for realistic recognition tasks. Simulation results well validate that the algorithmic power of deep multi-layer learning can be seamlessly merged with the efficiency of time-based spiking neuromorphic architecture, demonstrating great potentials of ``MT-Spike'' in resource and power constrained embedded platforms.

\end{abstract}


\input{intro}

\input{prelim}

\input{design}

\input{evaluation}
\input{conclusion}

\bibliographystyle{IEEEtran}
\bibliography{cites,NN,spiking} 

\end{document}

%% file: intro.tex
\section{Introduction}
\label{sec:intro}

The last decade has witnessed unprecedented evolutions of artificial intelligence (AI), since the deep learning systems such as deep neural networks (DNNs) and convolutional neural networks (CNNs) are developed to perform a series of human-level cognitive applications~\cite{lecun2015deep}. 
However, the underlying enormous computation and storage requirements seriously challenge DNNs' processing efficiency, and hence make them less attractive for cognitive tasks executing in many light-weighted platforms such as smart phone, wearable device and Internet-of-Things (IoT) etc., where very tighten power and hardware resources are enforced\cite{andri2016yodann,han2016mcdnn}.

Recently, spiking-based neuromorphic computing inspired by Spiking Neural Network (SNN), which is often recognized as the third-generation neural network that can closely embrace the working mechanism and efficiency of human brain, has emerged for achieving tremendous computing efficiency at much lower power of a single chip, i.e. total 1 million synapses with an operating power of ~70mW in IBM TrueNorth~\cite{akopyan2015truenorth}. To mimic the brain-style information processing, the input data of SNN is usually conveyed as the electrical spike train (or voltage pulse vector), followed by a more energy-efficient event-driven computation~\cite{Neil-FPGA-Spike2014}, thus it is a promising solution for hardware-favorable cognitive applications~\cite{neil2014minitaur,corradi2015neuromorphic}.

Similar as state-of-the-art DNNs or CNNs, an efficient multilayer learning rule to support the multilayer SNN architecture will be essential to enhance SNN's capability in realistic cognitive tasks. Many multilayer rate-based SNNs (rSNNs) are successfully prototyped to fulfill the real-world tasks~\cite{Esser11102016,akopyan2015truenorth,neil2014minitaur,corradi2015neuromorphic,liu2016memristor,seo201145nm,diehl2015fast,cao2015spiking} by directly borrowing the Backpropogation (BP) algorithm of Artificial Neural Network (ANN)~\cite{rumelhart1988learning}, as such a rate-based information representation is analogous to the numerical representation in ANN. 
However, the efficiency of rSNN largely relies on the number of spikes -- a large time window should be maintained for generating a huge number of spikes, resulting in an inefficient data processing and considerable spiking power consumption. On the other hand, time-based SNN (tSNN) can express the information more flexibly based on the presence and the delay of each generated spike. Moreover, a better energy-efficiency can be achieved by tSNN if the information can be efficiently embedded in extremely sparse spike trains, i.e. a single spike~\cite{bohte2002error}. However, unlike the rSNN, the realistic application of tSNN systems is still limited due to its weak learning capability. Developing efficient multilayer learning algorithms to enhance the potentials of tSNN is non-trivial due to its fundamentally different processing paradigm -- the time-based spiking voltage modulation with a non-differentiable thresholding function~\cite{gerstner2001framework,bohte2002error,mostafa2016supervised,xie2016efficient,zenke2017superspike,liu2017fast,shrestha2017stable}. Despite of many existing time-based learning rules like ``Tempotron''~\cite{gutig2006tempotron} and ``SpikeProp''~\cite{bohte2002error}, those proof-of-concept algorithms are neither compatible with multilayer extension nor feasible to handle the realistic applications due to theoretical limitations or expensive convergence of learning etc. Thus, an efficient multi-layer time-based learning algorithm that can merge the algorithmic power of deep learning to the efficiency of the time-based SNN architecture will be very crucial.

In this work, by orchestrating our proposed time-based coding and multi-layer learning algorithm, a \textit{\underline{M}ultilayer \underline{T}ime-based \underline{Spiking} Neuromorphic Architecture}, namely ``\textit{MT-Spike}", is proposed to facilitate the realistic cognitive applications. Our major contributions include: 
\begin{itemize}
\item We proposed a practical time-coding scheme to efficiently translate various types of information into the time domain through an individual spike, achieving remarkable reduction on spiking energy consumption and network model size; 
\item We developed a novel average delay response model to simplify the expensive neural processing in tSNN and enable the multilayer extension, significantly enhancing the learning capacity of this single-spike-driven neuromorphic computing system; 
\item We proposed a heuristic loss function and integrated it with the derived temporal error backpropagation algorithm, leading to a more efficient multi-layer learning for tSNN. 
\end{itemize}
Our evaluations show that ``MT-Spike" can even achieve the accuracy comparable to that of CNN while still maintaining the energy and processing efficiencies of tSNN when handling realistic tasks like ``MNIST" dataset, demonstrating a very promising solution for emerging cognitive computing on resource-limited platforms.

%% file: prelim.tex
\section{Backgrounds and Motivations}
\label{sec:prelim}

\subsection{Basics of rSNN and tSNN}

The popular spiking neural network (SNN) architectures can be generally categorized as Rate-based SNN (rSNN) and Time-based SNN (tSNN), where ``rate-coding'' and ``time-coding'' schemes are adopted to encode the input data~\cite{borst1999information}, respectively. 

\begin{figure}[t]
\centering
\includegraphics[width=1\columnwidth]{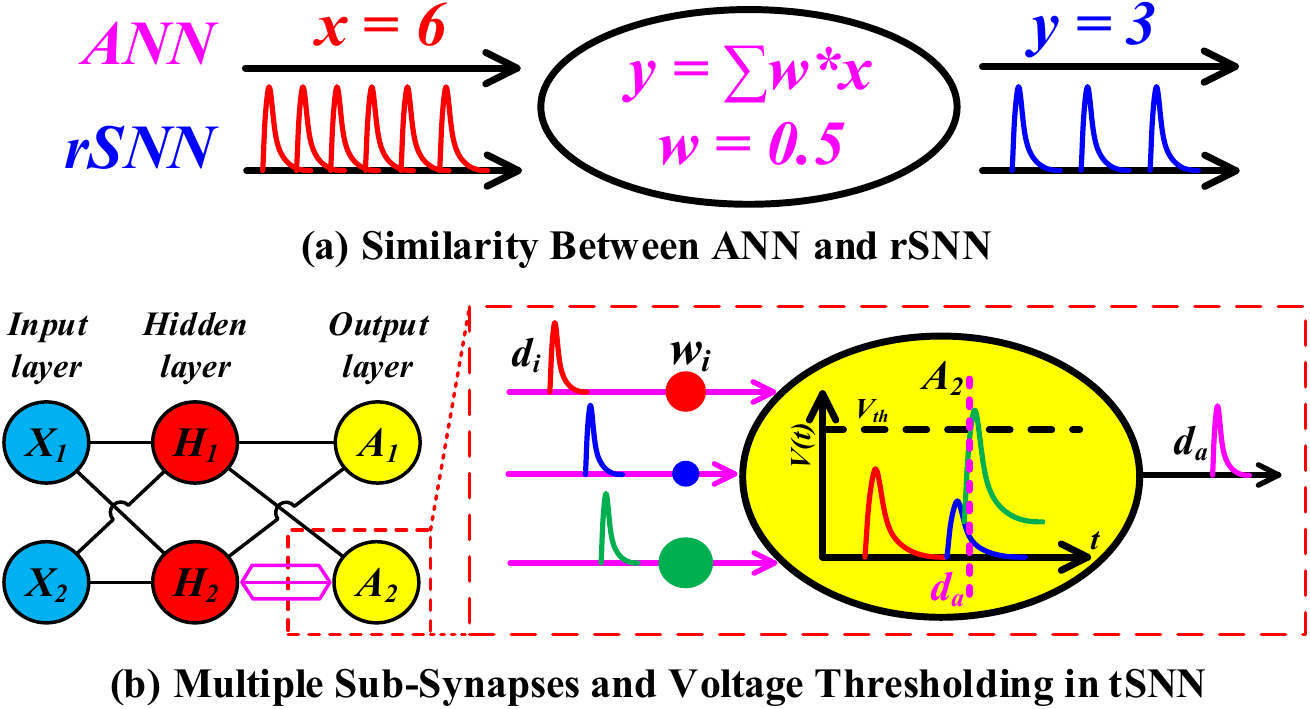}
\vspace{-5pt}
\caption{%
\label{prelim}
Neural processing in ANN, rSNN and tSNN.
}
\end{figure}

In rSNN, each piece of input information is first translated into a spike train of the input neuron with its occurrence frequency proportional to the numerical representation of the input data over a preset time period. For example, the number of spikes (i.e. ``6" here) in rSNN is equivalent to the intensity of input data (ANN-style, $x=6$), as Fig.~\ref{prelim}(a) shows. Then the spikes will be weighted towards the synthesized results of output neurons through the connected synapses. The patterns can be recognized based on response strength of the output neurons, e.g. the largest number of output spikes (or rate). Because the spike rate is closely analogous to information representation of the ANN, many practical multilayer rSNNs are well demonstrated in real-world applications by naturally adopting ANN's backpropogation (BP) algorithm. Moreover, the efficiency largely relies on the number of spikes because of such a rate-based information process mechanism~\cite{han2016energy}. 

The tSNN expresses the information more elaborately by leveraging both the presence and occurring time of individual spike, i.e. each stimulus is represented as the desired delay of a single spike in our design, thus ideally more energy-efficient than rSNN because of significant reduced number of spikes~\cite{gerstner2001framework,gerstner2008spike,burkitt2006review,borst1999information}. As Fig.~\ref{prelim}(b) shows, the input voltage pulses (kernel-modulated spikes) with different delays $d_i$ are tunned by synapses with different weights $w_i$ and then accumulated at the output neuron. Once the sum of membrane voltage reaches a target threshold, an output spike will be generated and the whole system can be stopped. Accordingly, its occurrence time $d_a$ can determine a data pattern.

\subsection{Impractical multi-layer learnings in tSNN}
\label{sec:motivation}
Extending the single-layer tSNN to multi-layer tSNN can potentially enhance its capability for realistic cognitive tasks. However, designing efficient tSNN multi-layer learning algorithms is very challenging due to the fundamentally different training mechanism---the time-based spiking voltage modulation with a non-differentiable thresholding function. We have investigated many existing time-based learning algorithms, i.e. 
unsupervised spiking-time-dependent plasticity (STDP)~\cite{sjostrom2010spike}, theoretical ``Tempotron'' learning~\cite{gutig2006tempotron} and ``SpikeProp''~\cite{bohte2002error}. Those proof-of-concept algorithms are either unable to support multi-layer structure or too bio-plausible to handle the realistic applications because of the cost and difficult convergence of learning etc.


Fig.~\ref{prelim}(b) illustrates the working principle of the most popular multi-layer supervised temporal learning algorithm- -``SpikeProp''~\cite{bohte2002error} in a two-layer tSNN. Here ``SpikeProp'' can perform complex nonlinear classification in temporal domain by customizing the BP algorithm widely adopted in multi-layer ANNs. Unlike the one-one synaptic connection of two neurons in a standard BP-based multi-layer ANN, the link between any two neurons of two adjacent layers in ``SpikeProp'' is composed of multiple synaptic terminals (i.e. $m$), where each terminal serves as a sub-synapse associated with a different spiking delay $d_i$ and weight $w_i$ (see the connection of example neurons $H_2$--$A_2$ in Fig.~\ref{prelim}(b)). A sufficient number of such sub-synapses that can precisely model small delay differences and modulate the spiking voltage kernels between each pre-synaptic and post-synatic neuron pair is needed, leading to significantly enlarged network size. As an example, handling the simple XOR problem with a two-layer architecture (one hidden layer and one output layer) requires $\sim40\times$ more weights in ``SpikeProp''-based tSNN~\cite{bohte2002error} than that of an ANN (240 v.s. 6). Thus, the limited scalability of such a bio-plausible algorithm greatly hinders it from solving more practical and complicated cognitive tasks regardless of the expensive implementation cost, e.g. accurately control the temporal information.

%% file: design.tex
\section{Design Details}
\label{sec:design}

In this section, we present the design details of our proposed ``MT-Spike" -- a multilayer time-based spiking neuromorphic architecture with temporal error backpropagation. 



\subsection{System Architecture}
As a realization of multilayer fully-connected spiking neural network (SNN), MT-Spike is inspired from biological spiking neuron models and able to work in ``training'' and ``testing'' modes for non-linear classification tasks. 
\begin{figure}[t]
\centering
\includegraphics[width=1\columnwidth]{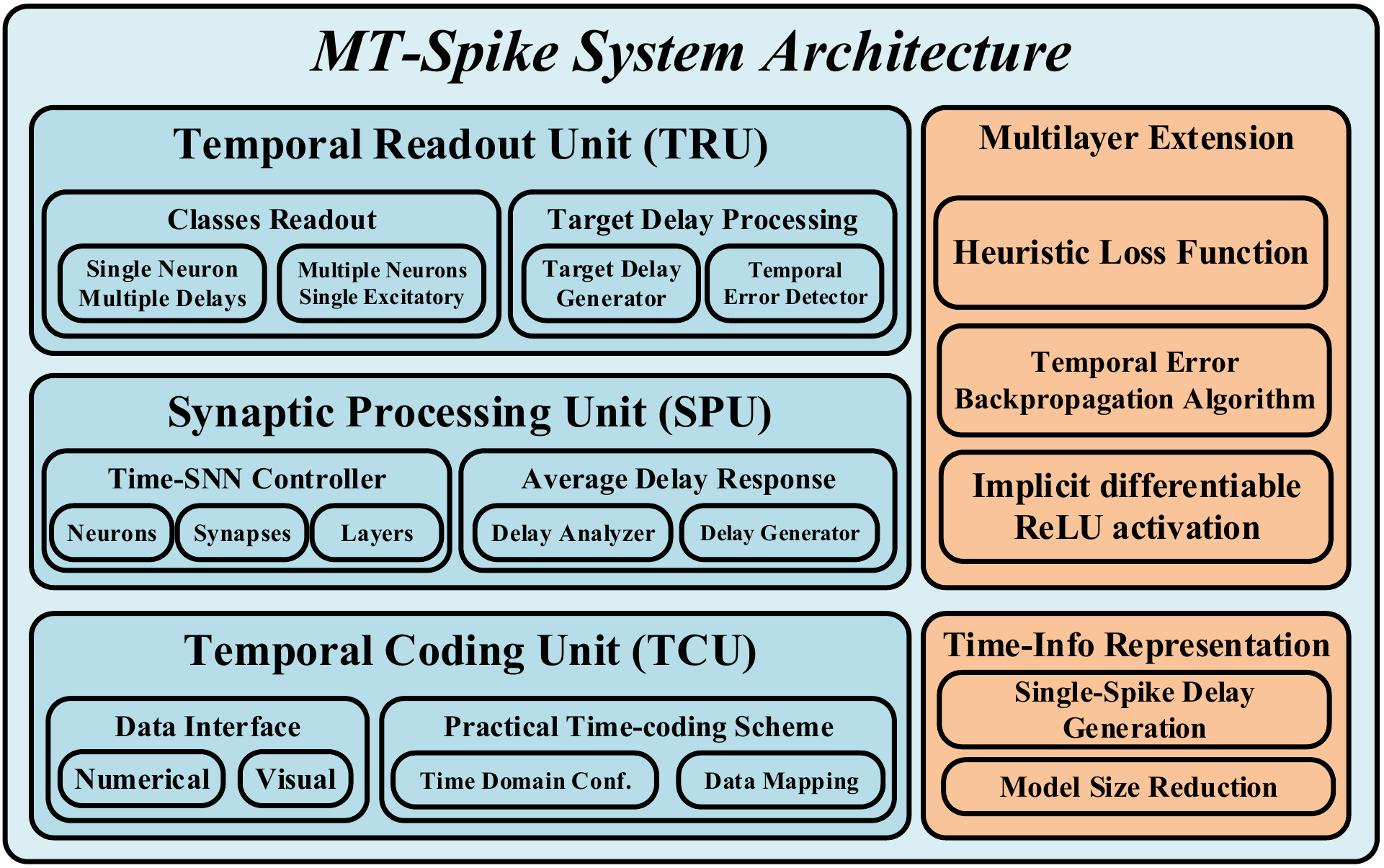}
\vspace{-5pt}
\caption{%
\label{arch}
The overview of MT-Spike system architecture.}
\vspace{-5pt}
\end{figure}
As Fig.~\ref{arch} shows, neural processing is conducted in MT-Spike through three major components -- Temporal Coding Unit (TCU), Synaptic Processing Unit (SPU) and Temporal Readout Unit (TPU).
\subsubsection{Temporal Coding Unit (TCU)}
TCU is developed to handle a variety of stimuli like numerical and visual samples at the input layer. With the underlying practical time-coding scheme, stimuli can be first translated into spike delays needed by each input neuron, then a time-based sparse spike train--single spike per input neuron will be generated and sent to Synaptic Processing Unit. Specially, a flexible neural network size reduction based on the temporal-spatial information conversion can be achieved by the proposed time-coding scheme. 
\subsubsection{Synaptic Processing Unit (SPU)}
As the major part of SNN, SPU consists of synapses and neurons which are organized in multiple layers. In each layer, the temporal information (i.e. delays), rather than the voltage kernels, of the coming spike train will be directly adjusted through synapses and integrated by neurons. 
With devised average delay response model, each neuron can obtain the customized temporal information and then immediately generate an output spike according to the calculated delays. The output spike train will be further sent to next layer following a similar processing mode until reaching the output layer. Note traditional tSNN tunes the voltage modulation based on the pre-synatic and post-synatic delay differences, fires a spike until the accumulated voltage reaches the threshold voltage and records the associated spike delay. However, SPU directly leverages the delays for fast computations and completely eliminates the costly and time-consuming spiking-kernel (voltage) related operations. 
\subsubsection{Temporal Readout Unit (TRU)}
TRU is responsible to perform the classification by directly reading out the delays of the final output spikes from the SPU. In training mode, the individual target spiking delay of each output neuron will be set by TRU and compared with the actual output delay for the temporal error detection and calibration. Through heuristic loss function and efficient temporal error backpropagation algorithm, only associated temporal errors from the output layer will be calculated and layer-wise back-propagated to update those correlated synapses.

\begin{figure}[t]
\centering
\includegraphics[width=1\columnwidth]{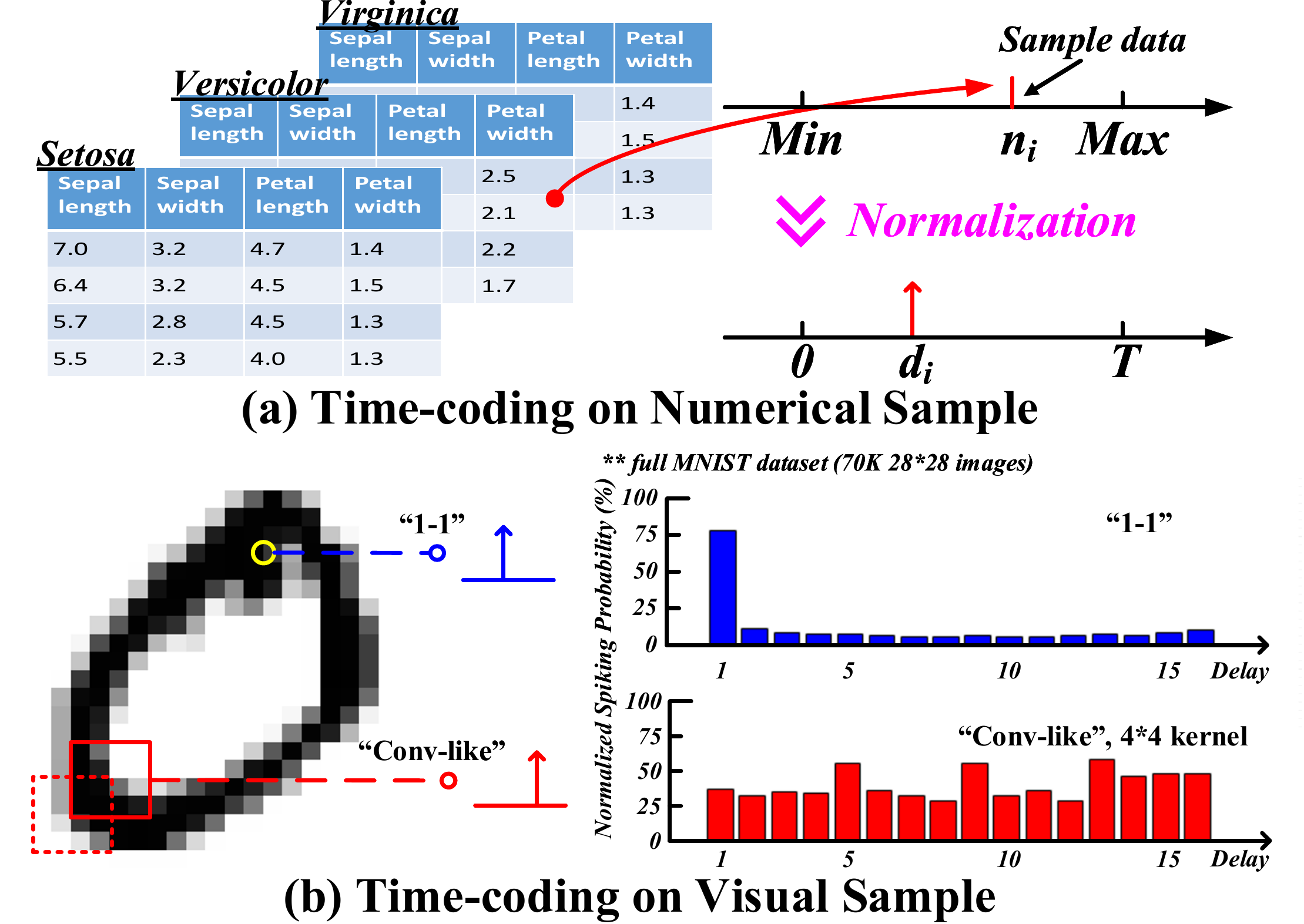}
\vspace{-10pt}
\caption{
\label{coding}
Practical Time-coding Scheme in MT-Spike.}
\end{figure}

\subsection{TCU and Practical Time-Coding}
As discussed previously, rSNN demands for a large number of spikes occurring in an adequate time window to represent the amplitude of input data (i.e. numerical value or pixel density). Because the information is only conveyed by the spiking rate, the additional coding dimension--the spike occurrence time in temporal domain, is not fully utilized for energy and processing efficiency optimizations. Hence, 
we propose ``practical time-coding scheme'' to efficiently link the input information to the occurrence time of generated spikes in TCU.
In our design, the input data will be carried by an ultra-sparse spike train -- a single spike per neuron with the information coded as the spiking delay, potentially suppressing the number of needed spikes towards better efficiency. 


\subsubsection{Practical Time-coding Scheme}
To better illustrate the proposed coding techniques, we define following three key parameters:  
An encoding time window $T$, a unit time interval $\tau$, and the time encoding resolution $R = \frac{T}{\tau}$. Note $\tau$ also denotes the period of a single spike. 
To make our encoding biological compatible, we also interpret the spike with a short (long) delay as the excitatory (inhibitory) response under strong (weak) stimulus. 

We explored several possible time-coding schemes on two representative datasets: numerical-style ``Iris dataset" (3 classes, 4 attributes)~\cite{fisher1936iris} and visual-style ``MNIST dataset" (10 handwritten digits)~\cite{lecun1998mnist}, as shown in Fig.~\ref{coding}.
In Iris dataset, each attribute (i.e. \{length, width ...\}) can be mapped to a single spike associated with an input neuron. As Fig.~\ref{coding}(a) shows, the delay $d_i$ of each single spike generated within $T$ can be calculated as $d_i = T\cdot round\left(1-\frac{n_i}{max{(n_i) - min(n_i)}}\right)$, where $n_i$ is the $i$-th data sample at a selected attribute. 
For visual-style ``MNIST dataset", we first investigated an existing coding technique adopted in most ANNs and SNNs -- the ``1-1 coding", i.e. each single pixel is mapped to an input neuron, as shown in Fig.~\ref{coding}(b). The delay $d_i$ of the spike generated by the input neuron $i$ is inversely proportional to the associated pixel density $p_i$: $d_i = T\cdot round\left(1-\frac{p_i}{max(p_i)}\right)$. Note there will be no spike if $p_i = 0$. However, the coding efficiency of ``1-1 coding" is limited because many spikes that should represent different data patterns occur at a common time slot (see the spiking delay distribution of ``1-1 coding'' in Fig.~\ref{coding}(b)). Besides, the number of input neurons is always equal to the image resolution, indicating a large model size. To better leverage the whole encoding time window and reduce the model size, we further develop the ``conv-like coding'' inspired by human visual cortex (receptive field) and Convolutional Neural Networks (CNNs). By perceiving the localized information from multiple adjacent pixels through a square kernel, spiking delay in ``conv-like coding'' can be expressed as the number of ``0s'' within the kernel among the binarized pixels. As Fig.~\ref{coding}(b) shows, the spiking delays of ``conv-like coding'' are almost evenly distributed across the whole time domain, indicating effective utilization of temporal information, thus a potential model size reduction in spatial domain or rather a reduced number of input neurons.

\begin{figure}[t]
\centering
\includegraphics[width=1\columnwidth]{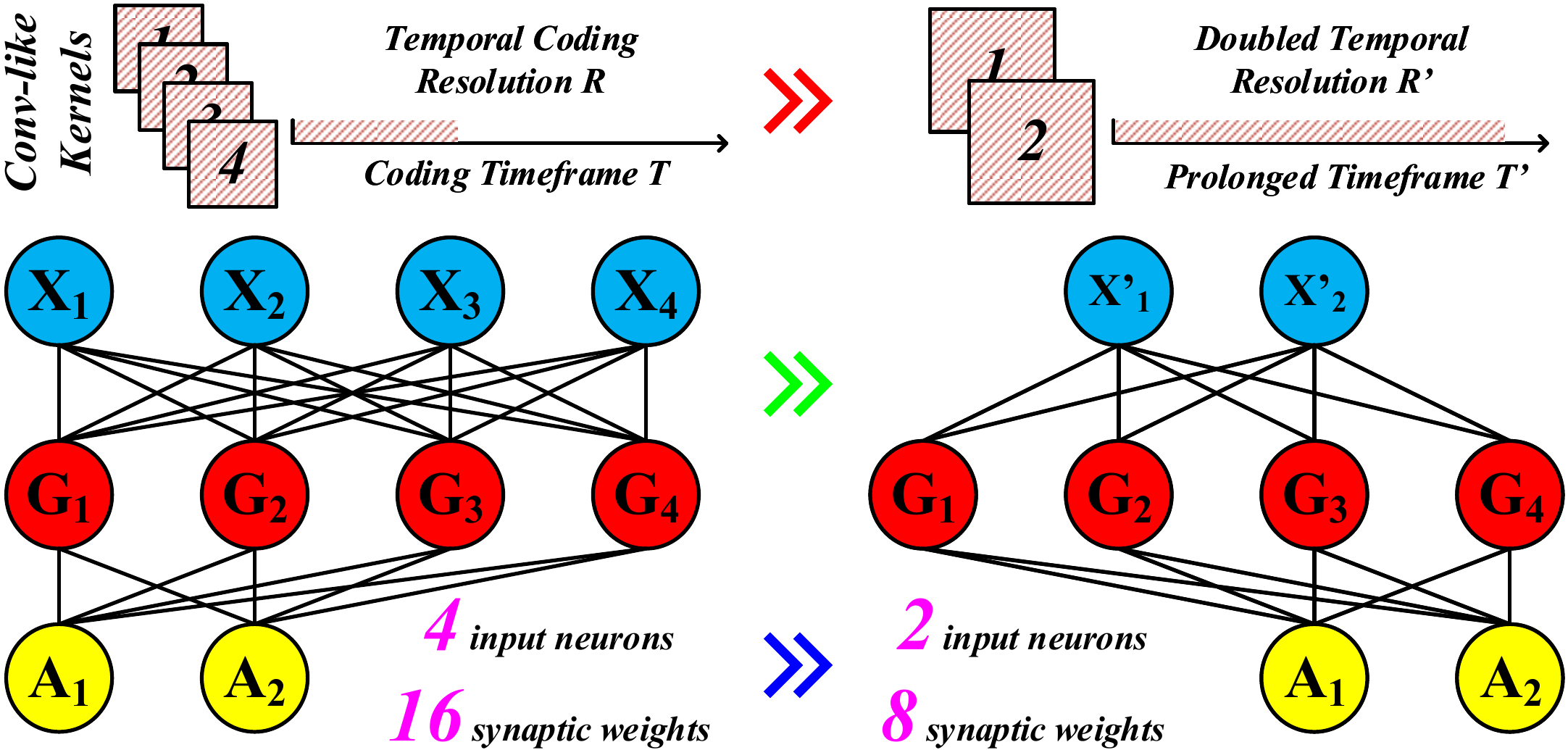}
\vspace{-5pt}
\caption{
\label{modelsize}
Model size reduction through adjustable temporal resolution.}
\end{figure}

\begin{figure*}[t]
\centering
\includegraphics[width=1\textwidth]{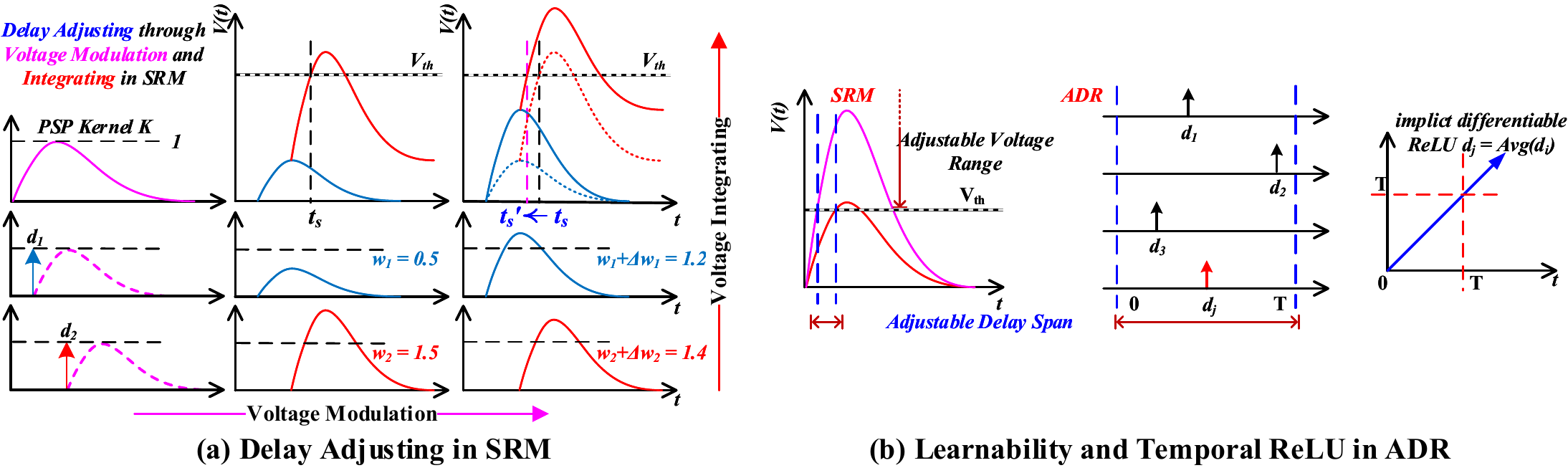}
\vspace{-10pt}
\caption{
\label{adr}
Design exploration on Average Delay Response Model.}
\end{figure*}

\subsubsection{Spatial Model Size Reduction}
\label{model_red}
To illustrate the advantage of spatial model size reduction provided by our proposed ``conv-like coding'', we assume the number of elements covered by the kernel as a square number $R$. Note $R = \frac{T}{\tau}$ also represents the temporal resolution of encoding. The number of input neurons can be expressed as $M = \lceil\frac{P-\sqrt{R}+1}{S}\rceil^{2}$, where $P$ and $S$ represent the width of input image and the stride to slide the kernel. ``Zero-padding" will be also applied according to the image resolution. Hence the encoding time window $T$ and input neuron number $M$ can be flexibly changed by tunning $R$ without sacrificing the amount of information of an entire image. Fig.~\ref{modelsize} shows the concept of model size reduction based on ``conv-like coding''. In this example, the ``original design'' is configured as $M=4$ input neurons, 16 synaptic weights for the first layer at a temporal resolution $R$. Alternatively, a ``size-reduced design" with only $M=2$ input neurons, 8 synaptic weights (50\% less), can be easily achieved by doubling the temporal resolution $R$ or rather the encoding time $T$ (assume $\tau$ does not vary). Although the efficiency of model size reduction depends on the percentage of the first-layer weights over the total number of weights, as we shall show later, such a technique is still very effective even without degrading the system accuracy.  




\subsection{SPU and Average Delay Response}
After the information is encoded as the delay of the input spike, the next question becomes how to perform the layer-wise time-based synaptic processing. The objective of the synaptic processing unit (SPU) is to generate an output response at each neuron based on its afferent input delays. Thus, how the neural processing model handles the temporal information will directly impact the performance of SPU in ``MT-Spike''. As discussed in section~\ref{sec:motivation}, the existing multi-layer tSNN still depends on expensive voltage modulation and threshold based neural processing paradigm due to the absence of the proper loss function and differentiable activation function, significantly hindering its applicability in real-world cognitive tasks. 


To develop an efficient time-based neural processing, we first explored the processing mechanism of biological plausible Spike Response Model (SRM)~\cite{gerstner2001framework,gutig2006tempotron,bohte2002error}.

\subsubsection{Delay Adjusting Through Weighting Efficacy}
Fig.~\ref{adr}(a) presents the concept of SRM. Its detailed  mathematical model can be expressed as:
\begin{equation}
\label{eq_srm_1}
\begin{cases}
V(t) = \sum_i w_i\sum_{d_i}K(t-d_i)\\
K(t-d_i) = \exp\left(-\frac{t-d_i}{\tau_1}\right) - \exp\left(-\frac{t-d_i}{\tau_2}\right)\\
V(t_s) = V_{th} \Rightarrow t_s = d_j
\end{cases}
\end{equation}
Where $K$, $\tau_1$ and $\tau_2$ are the Pre-Synaptic Potential (PSP) kernel function, voltage decay and integrate time constant, respectively. As Fig.~\ref{adr}(a) shows, the two updated weightings ($w_1+\Delta w_1$ and $w_2+\Delta w_2$) are applied to the two delayed versions ($d_1$ and $d_2$) of PSP spiking kernels, respectively. Accordingly, the integrated voltage w.r.t. time is slightly changed, translating into an equivalent delay adjustment when the voltage reaches the threshold ($t_s\to t_s^\prime$). Despite of the costly analog voltage computation and the target delay extraction, the fundamental goal of SRM is to identify an output spiking time by leveraging the pre-synaptic weights and input spiking delays.
Inspired by this observation, we propose the following Average Delay Response (ADR) Model (see Fig.~\ref{adr}(b)):
\begin{equation}
\label{eq_adr_1}
d_j(w_{ij},d_i)=\frac{1}{n}\sum_{i=1}^n w_{ij} d_i
\end{equation}
where $w_{ij}$, $d_i$ and $n$ denote the synaptic weighting efficacies between neuron $i$ and $j$, input spike delays of neuron $i$ and number of post-synapses. $d_j$ denotes the output spike delay of neuron $j$. Hence, the output spiking delay can be directly tuned by the weights $w_{ij}$, speeding up or slowing down the occurrence of an output spike. Note the result of ADR model (see Eq.~\ref{eq_adr_1}) is no less than any input delay $d_i$, which well complies with the nature of a causal system--a post-synaptic spike will be only trigged by the pre-synaptic input spikes. 

\subsubsection{Advantages of Average Delay Response Model}
First, the proposed ADR model can eliminate the costly voltage kernel modulations and complicated pre-synaptic/post-synaptic time control unavoidable in traditional tSNNs, because the proposed time-coding schemes ensure a comprehensive precise delay based information process across all the layers, e.g. performing target classification and error calculation by the delay.

Second, ADR model also increases the adjustable delay range significantly (e.g. a whole encoding time window $T$) by direct delay weighting when compared with that of traditional SRM limited by the PSP kernel, as shown in Fig.~\ref{adr}(b). As we will show in Section.~\ref{sec:evaluation}, ``MT-Spike" with average delay response can achieve remarkable improvement accuracy over the traditional tSNN.

Finally, ADR model can implicitly work as a ``Special ReLU"~\cite{nair2010rectified} function--a non-negative output delay with a smaller value representing a stronger response for an output neuron (the earlier the spike fires, the stronger the response is). Unlike the un-differentiable threshold function in traditional tSNN, the ``Special ReLU" function is differentiable and thus can facilitate an efficient multilayer learning through temporal error propagation.



\begin{figure*}[t]
\centering
\includegraphics[width=0.9\textwidth]{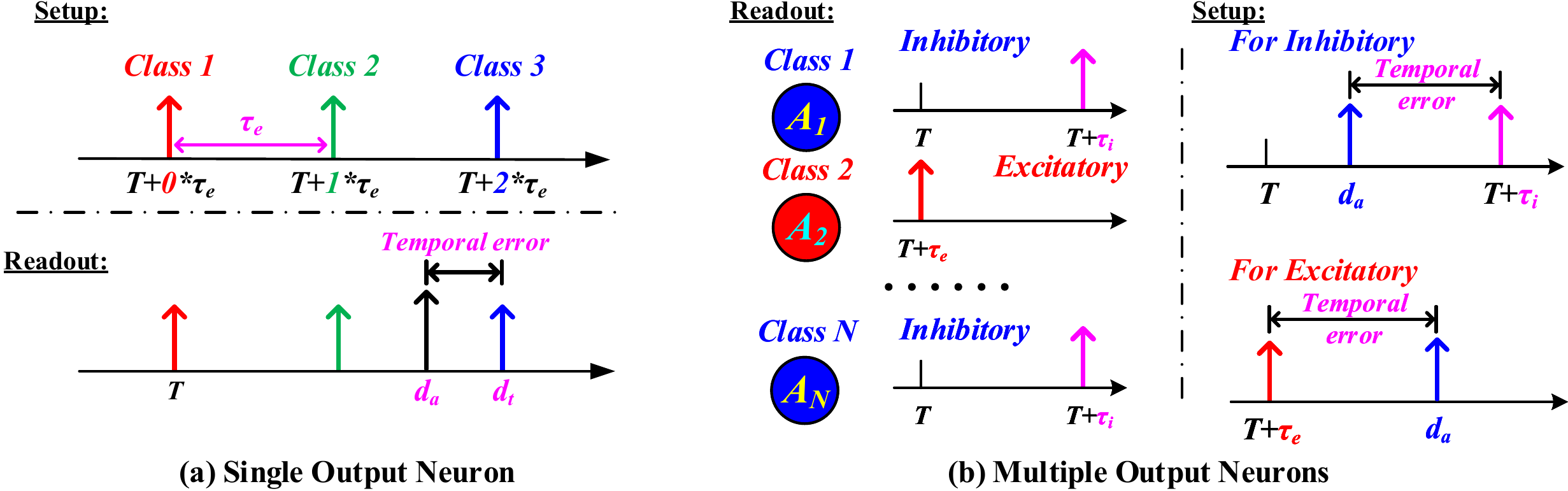}
\vspace{-5pt}
\caption{
\label{readout}
Target delay setup and class readout.}
\vspace{-5pt}
\end{figure*}

\subsection{Target Delay Set and Class Readout in TRU}
The functionality of Temporal Readout Unit (TRU) mainly consists of target delay setup and class readout for the testing and training modes. Since the practical time coding schemes can perform the spatial-temporal information conversion (see section~\ref{model_red}), we present the implementation details of Target Delay Setup and Class Readout for following two different cases: 1) A single output neuron with multiple target spiking delays, and the class number is equal to that of target delays; 2) Multiple output neurons with only two target delays, where the output neuron number and the number of classes are identical. Similar to the traditional bio-plausible tSNN~\cite{bohte2002error}, we assume the selected target delays are no less than the encoding time $T$ in ``MT-Spike''. 







\subsubsection{Single Output Neuron}
\label{son}
To maximize the temporal information of the output neuron while minimizing the number of output neurons, we assign multiple target spike delays at a single output neuron in ``MT-Spike'' (see Fig.~\ref{readout}(a)). Here one target delay represents one class, i.e. the target delay $T+i\times \tau_e$ for the $i$-th class, where $\tau_e$ is the adjustable time interval to differentiate two adjacent classes and is constrained as no less than $\tau$--the period of a single spike. For instance, the target delay can be defined as \{T, T+3, T+6\} for the three classes \{``Setosa'', ``Versicolour'' and ``Virginica''\} in ``Iris dataset''~\cite{fisher1936iris}, respectively. Here $\tau_e=3\tau,\tau=1$. 

As Fig.~\ref{readout}(a) shows, these target delays will serve as ``delay checkpoints'' to readout a class according to temporal distances between the actual output delay and those ``delay checkpoints'', that is, to find the nearest target with smallest temporal distance for a testing. During the training, a temporal error will be calculated based on the delay distance between actual delay and target delay of a class at output neuron if a classification failure happens.
However, as we shall show later in Section.~\ref{sec:evaluation}, such a single output neuron solution suffers from significant accuracy degradation on complex datasets with more class numbers, i.e. MNIST~\cite{lecun1998mnist}, because of very limited weighting effectiveness on single output neuron.



\subsubsection{Multiple Output Neurons}
\label{sec:mon}
To handle the large dataset with more classes, an alternative solution is to increase the number of output neurons, i.e. same as the number of classes, so that each class can be dedicated to one output neuron. To maintain the biological plausibility, short target delay $T+\tau_e$ will be only assigned to the ``excitatory'' output neuron (i.e. neuron $A_2$, representing current class label 2) while that of all the remained ``inhibitory'' neurons are assigned with a same longer delay $T+\tau_i$, as shown in Fig.~\ref{readout}(b). Here $\tau_e <\tau_i$. 

For example, if the target class label is ``1'' (i.e. handwritten digits from ``0'' to ``9'') in MNIST, ten target delays $\{T+4, T+0, T+4,...,T+4\}$ will be assigned to the ten output neurons $\{A_1, A_2, A_3,...,A_{10}\}$, respectively, Here we assume $\tau_e = 0$ and $\tau_i = 4$. During the testing, the class readout will be achieved by the ``excitatory'' output neuron with an ``earliest'' spike, i.e. the one with minimal actual spike delay. In training mode, each output neuron will calculate an individual temporal error based on the difference of its actual delay and target delay if an incorrect class label is identified.


\subsection{Temporal Error Backpropagation and Heuristic Loss Function}
Based on our proposed average response model and its implicit temporal ``ReLU" activation, an efficient multilayer learning algorithm can be obtained through temporal error backpropagation for ``MT-Spike". 


\subsubsection{Temporal Error Backpropagation}
In this section, we present our proposed temporal error backpropagation algorithm. For an output neuron j, the temporal error function is defined as:
\begin{equation}
\label{eq_err}
E_j = \frac{1}{2}\left(d_{t(j)}-d_{a(j)} \right)^2
\end{equation}
where $d_{t(j)}$ is its target delay and $d_{a(j)}$ is its actual delay, with implicit activation function $\varphi$, the output delay of neuron $j$ in layer $l$ is given as:
\begin{equation}
\label{eq_outneur}
d_j^l = 
\varphi({net}_j^l) = 
\varphi\left(\frac{1}{n}\sum\limits_{i=1}^n{w_{ij}^l}d_i^{l-1}\right)
\end{equation}
where $d_i^{l-1}$ is the pre-synaptic delay of the $i$-th neuron and $n$ is the number of pre-synapses. Thus the partial derivative of temporal error with respect to weight $w_{ij}^l$ can be expressed as:
\begin{equation}
\label{eq_deriv}
\frac{\partial E_j}{\partial w_{ij}^l} = 
\frac{\partial E_j}{\partial d_j^l}
\frac{\partial d_j^l}{\partial {net}_j^l}
\frac{\partial {net}_j^l}{\partial w_{ij}^l}
\end{equation}
where:
\begin{equation}
\label{eq_p1}
\frac{\partial {net}_j^l}{\partial w_{ij}^l}=
\frac{\partial }{\partial w_{ij}^l}\left(\frac{1}{n}\sum\limits_{i=1}^n{w_{ij}^l}d_i^{l-1}\right)=
\frac{d_i^{l-1}}{n}
\end{equation}
\begin{equation}
\label{eq_p2}
\frac{\partial d_j^l}{\partial {net}_j^l}=
\frac{\partial }{\partial {net}_j^l}\varphi\left({\rm{net}}_j^l\right)=1
\end{equation}
For neuron $j$ at output layer $l$:
\begin{equation}
\label{eq_p3}
\frac{\partial E_j}{\partial d_j^l}=
\frac{\partial E_j}{\partial d_{a(j)}}=
\frac{\partial }{\partial d_{a(j)}}\frac{1}{2}{(d_{t(j)}-d_{a(j)})^2}=
d_{a(j)}-d_{t(j)}
\end{equation}
\begin{equation}
\label{eq_simp}
\frac{\partial E_j}{\partial w_{ij}^l}=\frac{d_i^{l-1}(d_{a(j)}-d_{t(j)})}{n}
\end{equation}
For neuron $j$ at hidden layer(s):
\begin{equation}
\label{eq_inner}
\frac{\partial E_j}{\partial d_j^l}=
\sum_{k=1}^n {\left({\frac{\partial E_j}{\partial {net_k^{l+1}}}\frac{{\partial {net_k^{l+1}}}}{\partial d_j^l}} \right)}=
\sum_{k=1}^n {\left({\frac{\partial E_j}{\partial d_j^l}\frac{\partial d_j^l}{\partial {net_k^{l+1}}}w_{jk}^{l+1}} \right)} 
\end{equation}
where $k$ is the post-synaptic neuron of $j$, by defining:
\begin{equation}
\label{eq_theta}
{\delta_j^l}=
\frac{\partial E_j}{\partial d_j^l}
\frac{\partial d_j^l}{\partial {net}_j^l}=
\begin{cases}
d_{a(j)}-d_{t(j)} &,l\ is\ output\ layer\\
\sum_k\delta_k^{l+1}w_{jk}^{l+1} &,l\ is\ hidden\ layer
\end{cases}
\end{equation}
We can obtain the weight updating at learning rate $\eta$ as:
\begin{equation}
\label{eq_update}
\Delta w_{ij}^l =  -\eta \frac{\partial E_j}{\partial w_{ij}^l}=
-\eta \delta_j^l\frac{d_i^{l-1}}{n}
\end{equation}

\subsubsection{Heuristic Loss Function}

\begin{figure}[t]
\centering
\includegraphics[width=1\columnwidth]{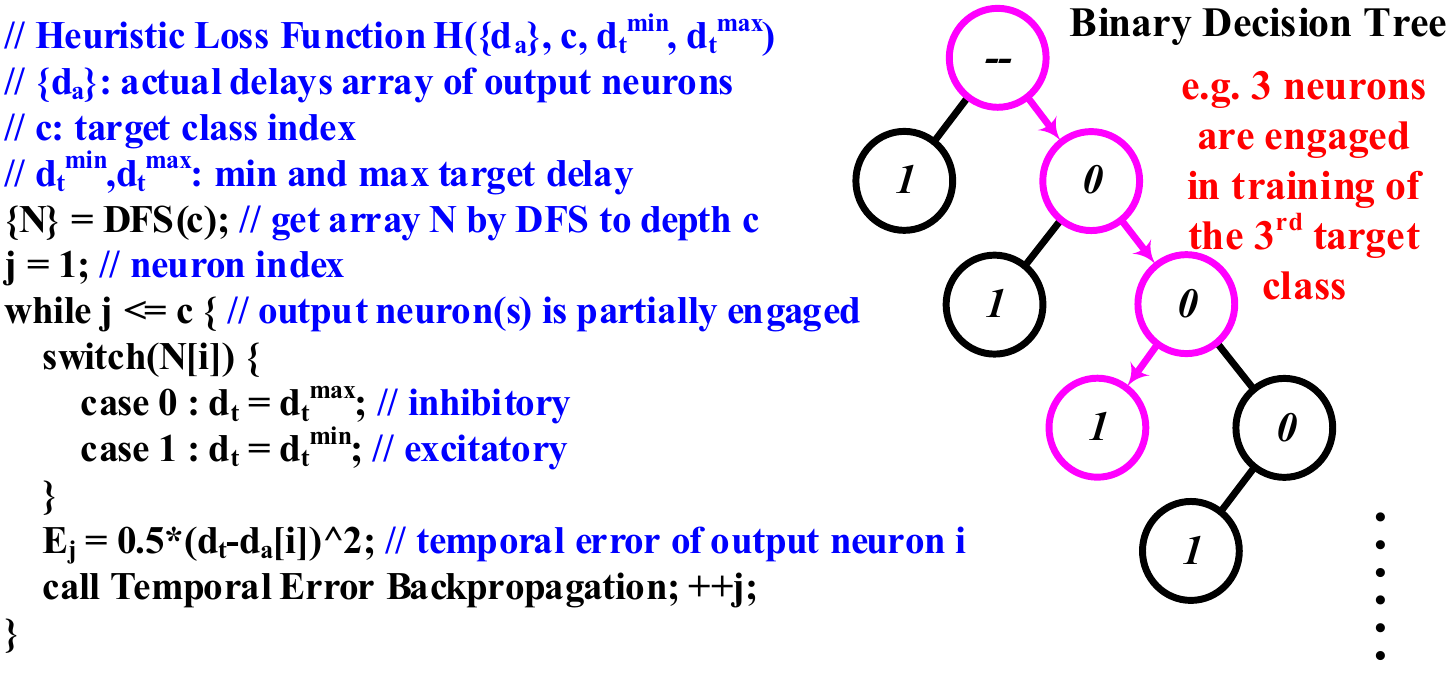}
\vspace{-10pt}
\caption{
\label{heu}
Heuristic Loss Function and Binary Decision Tree.}
\vspace{-5pt}
\end{figure}

In MT-Spike, the neural competition among different data patterns increases significantly as the dataset becomes more complicated, as the weight updating solely relies on the extreme sparse spike--single spike. As we will show later in Section.~\ref{sec:evaluation}, our MT-Spike exhibits lower accuracy compared with the multi-layer ANN when handling large complex dataset. Hence, to alleviate the neural competition, we further propose the Heuristic Loss Function in MT-Spike as the trigger of Temporal Error Backpropagation-- $H(\{d_a\},c,d_t^{min},d_t^{max})$, where $\{d_a\}$ and $c$ 
are the actual delay array of all output neurons and active class of current sample, respectively. $d_t^{min}$ and $d_t^{max}$ represent two target delays for excitatory neuron and inhibitory neuron, respectively (see section.~\ref{sec:mon}).


Fig.~\ref{heu} illustrates the algorithm, as well as the novel data structure of heuristic loss function. An ``Huffman'' style binary decision tree with its depth equal to the total number of target classes is introduced. Only partial output neuron(s) will be involved by leveraging a depth-first-search (DFS) through the binary decision tree. For example, to process the MNIST dataset (10 classes with label ``0'' to ``9''), the binary decision tree with a maximum depth $10$ (the depth of the root is $0$) will be generated according to Fig.~\ref{heu}. All the nodes, except the root node, in the left (right) subtree are marked as $1$ ($0$). If the $3^{rd}$ data pattern (class label ``2'') is selected, a depth-first-search will be conducted on the decision tree until the depth reaches $3$. The $3$ nodes traversed by the longest searching path (highlighted in Fig.~\ref{heu}) indicate only $3$ out of total $10$ neurons, i.e. $A_1$ and $A_2$ as inhibitory neuron and $A_3$ as excitatory neuron, will participate in the learning of the class ``2''. Note here only the synaptic weights associated with those three neurons will be updated. 
By deploying the Heuristic Loss Function in temporal error backpropagation of ``MT-Spike", the computation of the error $\delta$ (see Equation.~\ref{eq_theta}) can be further simplified as: 
\begin{equation}
\label{eq_theta_heu}
\begin{cases}
\delta_{j\in\Gamma}^l = d_{a({j\in\Gamma})}-d_{t({j\in\Gamma})} &,output\ layer\\
\delta_j^l=\sum_{k\in\Gamma}\delta_{k}^{l+1}w_{jk}^{l+1} &,last\ hidden\ layer\\
\delta_j^l=\sum_k\delta_k^{l+1}w_{jk}^{l+1} &,other\ hidden\ layer
\end{cases}
\end{equation}
where $\Gamma$ is the set of involved neurons, rather than the whole neurons, for a certain data pattern. In output layer, the weight updating will be partially conducted on the pre-synaptic weights of participated neuron(s):
\begin{equation}
\label{eq_update_heu}
\Delta w_{i(j\in\Gamma)}^l =  -\eta \frac{\partial E_{j\in\Gamma}}{\partial w_{ij\in\Gamma)}^l}=
-\eta\delta_{j\in\Gamma}^l\frac{d_i^{l-1}}{n}
\end{equation}
Such a pattern dependent partial weights updating rule can significantly reduce the weights competition, thus to boost the accuracy of ``MT-Spike'', as we shall show later.

\begin{table*}[t]
\centering
\caption{\label{tbl_sp}Structural Parameters of Selected Candidates.}
\resizebox{0.75\textwidth}{!}{%
\begin{tabular}{|c|c|c|c|c|c|}
\hline
Candidate & Types & Dataset & \begin{tabular}[c]{@{}c@{}}Network \\ Structure\end{tabular} & \begin{tabular}[c]{@{}c@{}}Number of \\ synaptic weights\end{tabular} & \begin{tabular}[c]{@{}c@{}}Neural processing \\ time-frame T\end{tabular} \\ \hline\hline

MT-1 & tSNN & Iris & 4-25-1 & 125 & 16+6 \\ \hline
SLMT-3 & tSNN & Iris & 4-3 & 12 & 16+4 \\ \hline
SpikeProp & tSNN & Iris & 4-25-3 & 3500 & 16+4 \\ \hline
MLP & ANN & Iris & 4-25-3 & 175 & -- \\ \hline\hline

MT-1 & tSNN & MNIST & 169-500-1 & 85000 & 16+9 ($\tau=0.1$) \\ \hline
MT-10(heu/noheu) & tSNN & MNIST & 169-500-10 & 89500 & 16+4 ($\tau=0.1$) \\ \hline
SLMT-10(heu/noheu) & tSNN & MNIST & 169-10 & 1690 & 16+4 ($\tau=0.1$) \\ \hline
SpikeProp & tSNN & MNIST & 784-500-10 & 7940000 & 16+4 ($\tau=0.1$) \\ \hline
Diehl & rSNN & MNIST & 784-6400 & 5017600 & 50 ($\tau=0.1$) \\ \hline
Minitaur & rSNN & MNIST & 784-500-500-10 & 647000 & -- \\ \hline
Lenet-5 & CNN & MNIST & 1024-C1-S2-C3-S4-C5-F6-10 & 60840 (340908 conn.) & -- \\ \hline

\end{tabular}%
}
\end{table*}

%% file: evaluation.tex
\section{Evaluations}
\label{sec:evaluation}


In this section, we will evaluate the accuracy, model size and power consumption of the proposed ``MT-Spike" architecture. Experiments are conducted in the platforms like MATLAB and heavily modified open-source simulator--Brian~\cite{goodman2009brian}.



\subsection{Experiment Setup}
Two representative datasets are selected as the benchmarks of our experiments, including ``Iris''~\cite{fisher1936iris} and ``MNIST''~\cite{lecun1998mnist}. ``Iris'' consists of 3 classes, with 50 samples per class and 4 numerical attributes per sample. Note the NOT-linear separable nature of the 3 classes can validate the functions of multilayer temporal-learning based ``MT-Spike", as well as its classes readout based on the multiple target delays of a single output neuron (see section~\ref{son}). We utilize 120 and 30 samples for training and testing purposes, respectively. The ``MNIST'' dataset, which includes 10 handwritten digits with 60K training images and 10k testing images, is adopted to evaluate the visual recognition capability of ``MT-Spike" in terms of accuracy, model size and approximated energy consumption. Several representative candidates, such as multi-layer ANNs, rSNNs and tSNNs, are implemented for a comparison purpose. Batch training is conducted in our evaluation. All the training samples are randomly fed into the candidates per epoch with a $batch size = 30$ ($256$) for ``Iris'' (``MNIST'') until the networks converge, followed by a testing iteration. Table.~\ref{tbl_sp} shows the detailed configurations and network types of all selected candidates. All ``MT-Spikes" are implemented with a same time window parameter $T = 16$ and learning rate $\eta = 0.01$. The initial weights $w\in\left(0,1\right)$ are randomly generated before training.

\subsection{Multilayer Validation on Iris Dataset}

\begin{figure}[b]
\centering
\includegraphics[width=1\columnwidth]{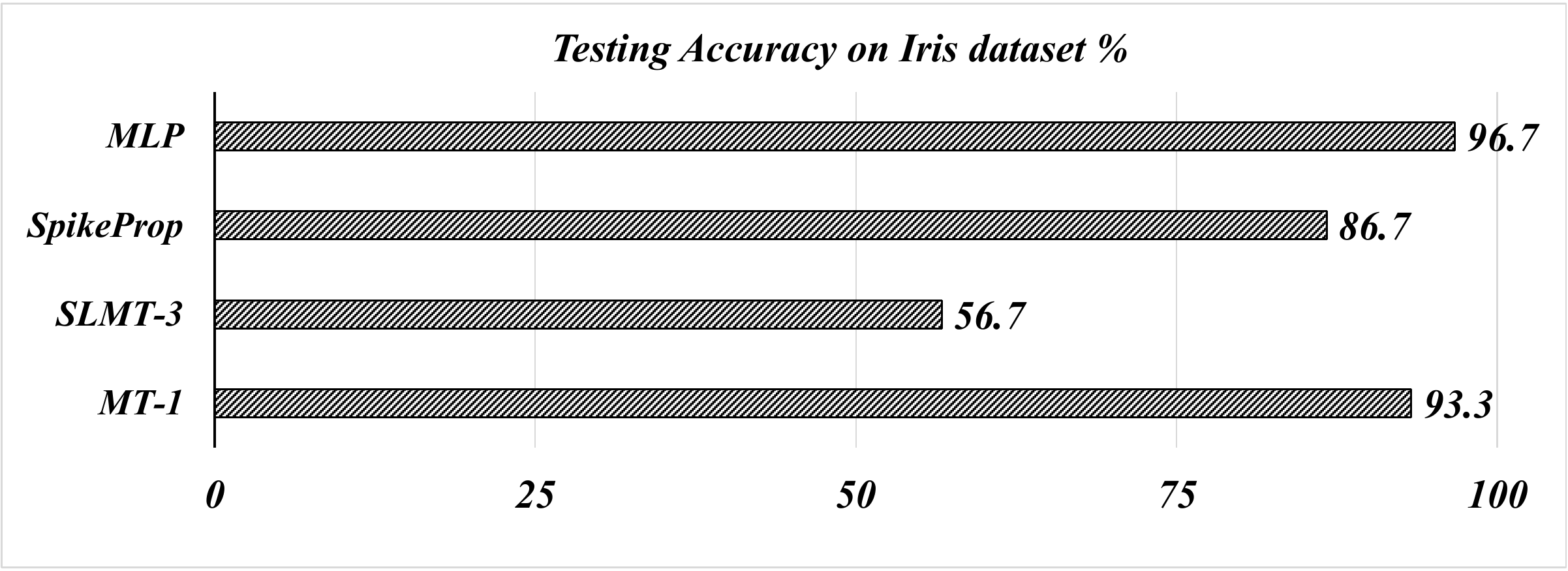}
\caption{
\label{e_iris}
Testing Accuracies on Iris Dataset.}
\end{figure}

As shown in Table.~\ref{tbl_sp}, ``Iris" dataset is used to evaluate the following four networks: ``MT-1''-- a multilayer MT-Spike implementation with only single output neuron and multiple target delays setup;``SLMT-3''-- A simplified version of MT-Spike without hidden layer; ``SpikeProp"--traditional bio-plausible multi-layer tSNN with voltage modulation and thresholding process~\cite{bohte2002error}; ``MLP''--A Multilayer Perceptron based ANN~\cite{rumelhart1985learning}.


Fig.~\ref{e_iris} compares the testing accuracy of the four aforementioned candidates. As expected, ``SLMT-3" exhibits the worst accuracy ($56.7\%$) among all candidates because this single-layer tSNN cannot well distinguish the NOT-linear separable classes. On the contrary, ``MT-1'' achieves much better accuracy than that of``SpikeProp" ($96.7\%$ v.s. $86.7\%$), and can even approach that of ``MLP'', demonstrating the enhanced capability through the proposed multi-layer temporal learning rule. Furthermore, as Table.~\ref{tbl_sp} shows, ``MT-1" reduces the synaptic weights by $\sim28\times$ compared with the ``SpikeProp'', which well validates the efficiency of single output neuron readout and the Average Delay Response model when handling the simple dataset.



\subsection{Performance Evaluation on MNIST Dataset}
To further evaluate the performance of our proposed ``MT-Spike" in a relative complicated dataset ``MNIST'', seven different networks with more network parameters are chosen, as shown in Table.~\ref{tbl_sp}. Here `` Diehl'' is an rSNN trained by the unsupervised STDP learning~\cite{diehl2015unsupervised}. ``Minitaur'' is a hardware-oriented rSNN towards power optimization. Besides, the CNN implementation -- ``Lenet-5" is included as well for a comparison purpose. For a fair comparison with other SNN candidates, the minimal time interval is set as $\tau = 0.1$ to provide a precise time-based processing for all ``MT-Spike" candidates.



\subsubsection{Model Size Reduction and Time-coding Efficiency}

\begin{figure}[b]
\centering
\includegraphics[width=1\columnwidth]{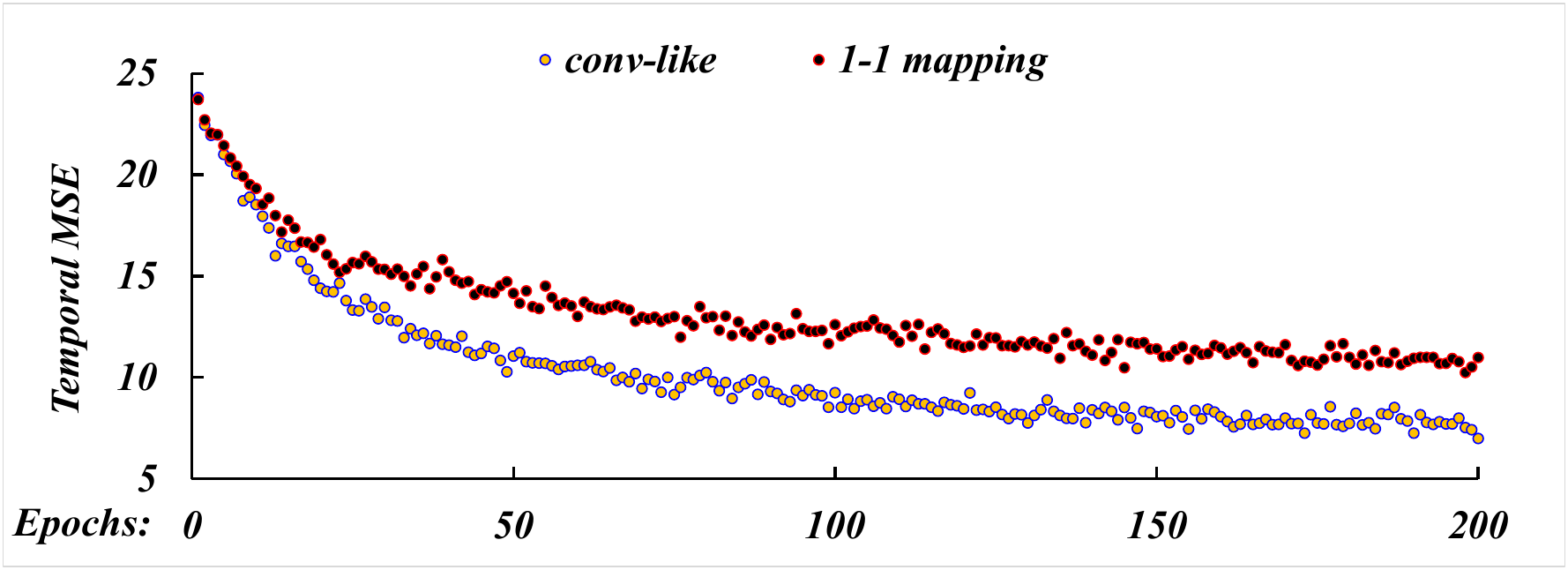}
\caption{
\label{e_coding}
Coding efficiencies of Conv-like and 1-1 mapping.}
\end{figure}

We first demonstrate the advantages of model size reduction in ``MT-Spike'' through the proposed ``conv-like'' time-coding scheme. As shown in Table.~\ref{tbl_sp}, the proposed ``MT-10" achieves $\sim4.6\times$ reduction on the number of input neurons (169 v.s.784) when compared with all the other non-``MT-Spike" candidates (except the ``Lenet-5" with 1024 neurons), which translates into an impressive model size reduction (or the number of weights) over ``SpikeProp", ``Diehl" and ``Minitaur", that is, $\sim88\times$, $\sim56\times$ and $\sim7\times$, , respectively. Note the ``SpikeProp" suffers from the largest model size due to a substantial number of sub-synapses between two connected neurons. As we shall discuss later, ``MT-10" can even maintain a very high accuracy despite of the significant reduced model size. 


Fig.~\ref{e_coding} also shows temporal mean-square error (MSE) v.s. training epoch for two ``MT-10'' designs that employ the ``conv-like'' coding and ``1-1 mapping'' coding, respectively. As Fig.~\ref{e_coding} shows, the adopted ``conv-like" coding achieves a lower MSE than that of ``1-1 mapping" coding at the same epoch, due to its better utilization of temporal information, e.g. the equally distributed spiking delays.


\subsubsection{Accuracy Analysis on MNIST dataset}

\begin{figure}[b]
\begin{centering}
\includegraphics[width=1\columnwidth]{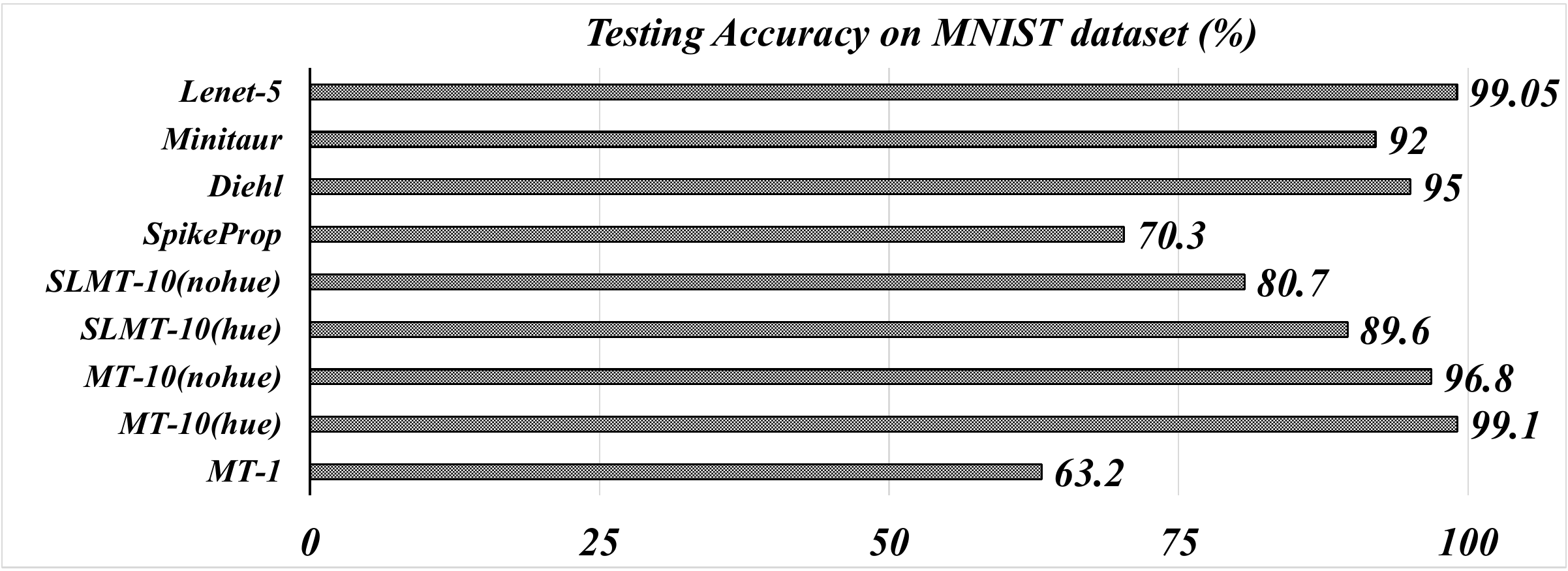}
\par\end{centering}
\caption{
\label{e_mnist}
Testing accuracy on MNIST dataset. 
}
\end{figure}

Fig.~\ref{e_mnist} shows the testing results of MNIST dataset among all different designs. As expected, ``MT-1" with single output neuron readout is insufficient to handle the complex dataset, resulting in the worst accuracy $63.2\%$, due to its weak weighting efficiency. 

We also evaluate the capability of the proposed heuristic loss function. As Fig.~\ref{e_mnist} shows, under a single-layered structure ``SLMT'', such a technique can boost the accuracy from $80.7\%$ on ``SLMT-10(nohue)" to $89.6\%$ on ``SLMT-10(hue)", showing a considerable accuracy improvement by alleviating the neural competitions. Moreover, by integrating the heuristic loss function with temporal error backpropagation, the accuracy of ``MT-10(hue)'' can be further increased to $99.1$\%, the best results among all candidates and even comparable with the CNN--``Lenet-5''($99.05$\%). Note the heuristic loss function can still introduce 2.3\% accuracy improvement in the multi-layer structure (``MT-10(hue)'' 99.1\% v.s.``MT-10(nohue)'' 96.8\%).

\subsubsection{Energy Consumption}

\begin{figure}[t]
\centering
\includegraphics[width=1\columnwidth]{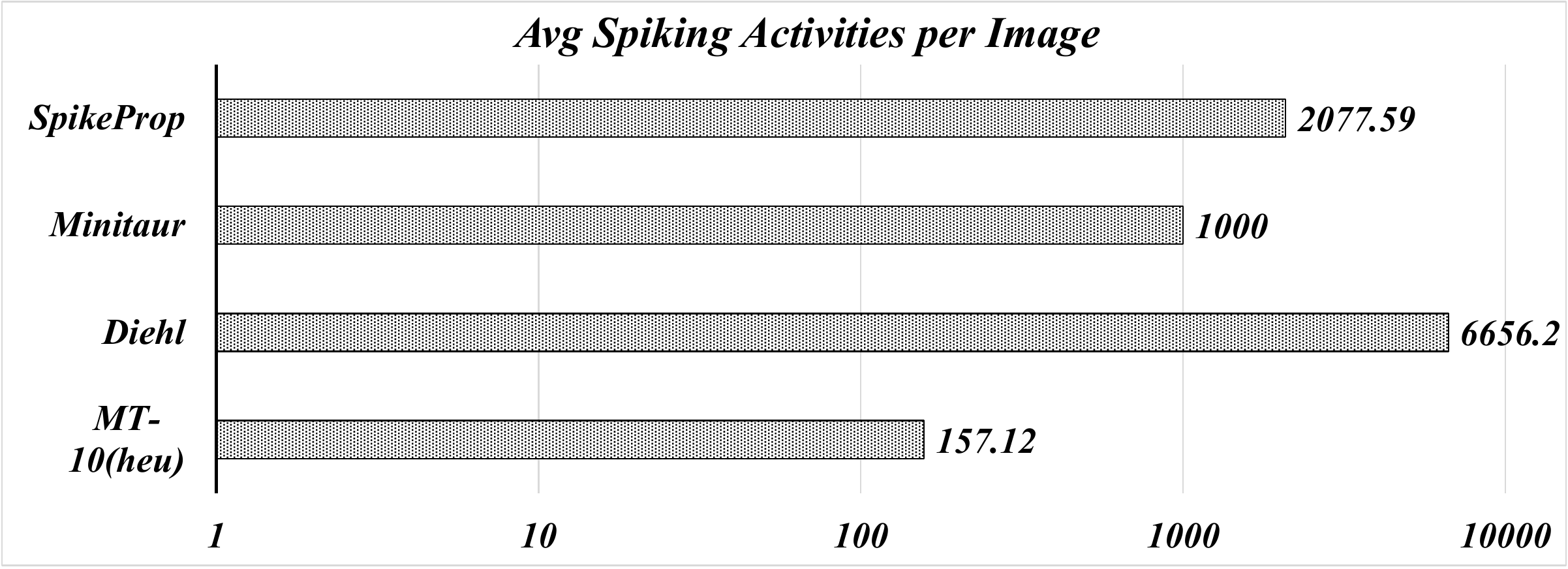}
\caption{
\label{e_power}
Energy analysis based on spiking activities. 
}
\end{figure}
To estimate the energy efficiency of ``MT-Spike", we adopt a similar estimation methodology presented in ~\cite{akopyan2015truenorth,cao2015spiking}. Measurement is conducted based on the following assumption: a single spike activity consumes $\alpha Joules$ of energy. The total spiking energy is calibrated based on the statistic of the spikes in testing iterations. As shown in Fig.~\ref{e_power}, ``MT-10(hue)" saves $\sim13\times$ power over ``SpikeProp", indicating the efficiency of our proposed average delay response model. Compared with rate-based ``Diehl", a $\sim42\times$ energy reduction is further achieved by ``MT-10(hue)" through the efficient single spike temporal representation. Moreover, ``MT-10(hue)" can still achieve $\sim6.3\times$ power reduction compared with the hardware-oriented design ``Minitaur", indicating an energy efficient solution for resource-limited embedded platforms.

%% file: conclusion.tex
\section{Conclusion}
\label{sec:conclusion}
Modern deep learning enabled neural networks are subject to great challenges on resource-limited platforms due to the enormous computation and storage requirements. Time-based spiking neural network (tSNN) has emerged as a promising solution, however, its capability of handling realistic tasks is significantly limited by the expensive biological plausible neural processing mechanism and theoretical time-based learning approaches, leading to inefficient information processing and impracticable multilayer-based deep learning. In this work, we propose a multilayer time-based spiking neuromorphic architecture, namely ``MT-Spike". Through a holistic solution set -- practical time-coding scheme, average delay response model, temporal error backpropagation algorithm and heuristic loss function, ``MT-Spike'' can deliver impressive learning capability while still maintaining its power-efficient information processing at a more compact neural network. Our evaluations well demonstrate the advantages of ``MT-Spike" over other rSNN and tSNN candidates in terms of accuracy, neural network model size and power.

\section{Acknowledgements}
This work is supported in part by NSF under project CNS-1423137, and the 2016-2017 Collaborative Seed Award Program of Florida Center for Cybersecurity (FC$^2$).